\def\year{2021}\relax
\documentclass[letterpaper]{article} 
\usepackage{aaai21}  
\usepackage{times}  
\usepackage{helvet} 
\usepackage{courier}  
\usepackage[hyphens]{url}  
\usepackage{graphicx} 
\usepackage{natbib}  
\usepackage{caption} 
\usepackage{graphicx}
\usepackage{comment}
\usepackage{amsmath,amssymb} 
\usepackage{color}
\usepackage{epsfig}
\usepackage{array}
\usepackage{diagbox}
\usepackage{booktabs}
\newcommand{\etal}{\textit{et al.}}
\usepackage[switch]{lineno}

\title{One-shot Face Reenactment Using Appearance Adaptive Normalization}

\author{
	Guangming Yao\textsuperscript{\rm 1}\thanks{Both authors contributed equally to this research.},
	Yi Yuan \textsuperscript{\rm 1}\thanks{Correspongding author},
	Tianjia Shao$^*$\textsuperscript{\rm 2},
	Shuang Li\textsuperscript{\rm 3},
	Shanqi Liu\textsuperscript{\rm 4},
	Yong Liu\textsuperscript{\rm 4},
	Mengmeng Wang\textsuperscript{\rm 4},
	Kun Zhou\textsuperscript{\rm 2}\\
}

\affiliations {
	\textsuperscript{\rm 1} NetEase Fuxi AI Lab \\
	\textsuperscript{\rm 2} State Key Lab of CAD\&CG, Zhejiang University \\
	\textsuperscript{\rm 3} School of Computer Science and Technology, Beijing Institute of Technology\\
	\textsuperscript{\rm 4} Institute of Cyber-Systems and Control, Zhejiang University\\
}

\begin{document}
	
	\maketitle
	\begin{figure*}[h]
		\centering
		\scalebox{1}[1]{
			\includegraphics[scale=0.2]{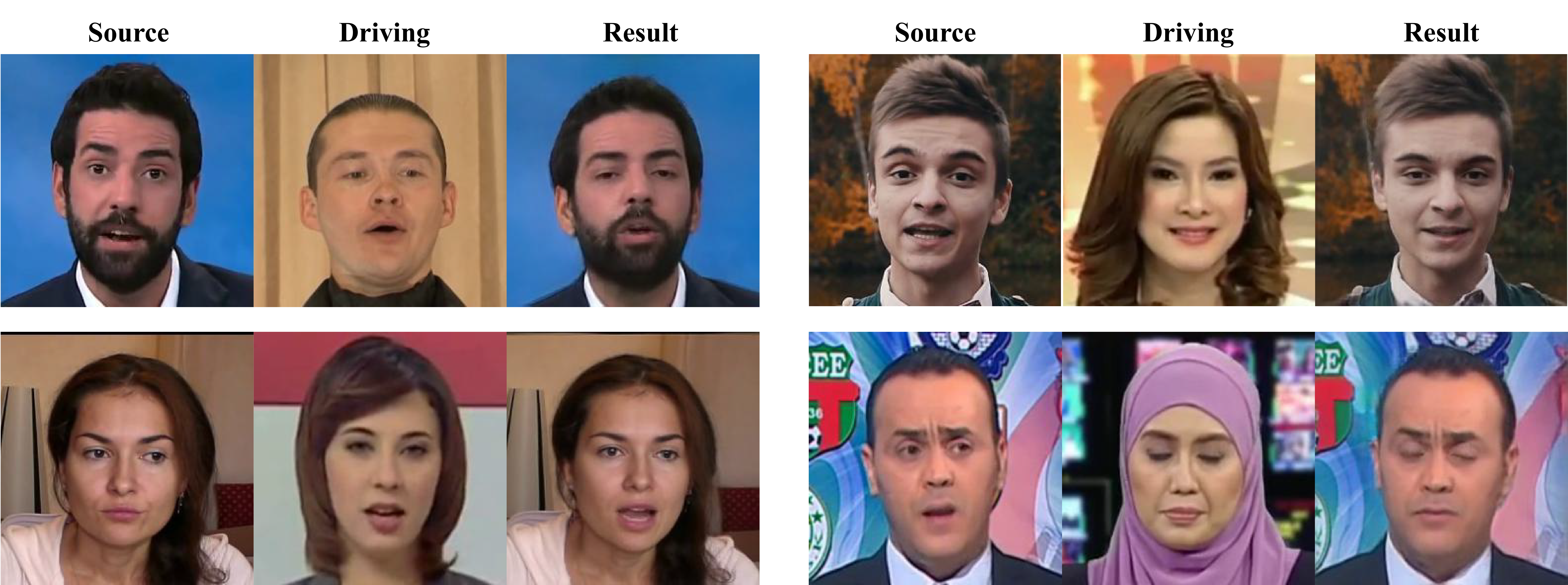} }
		\caption{Generated examples by our method. The source image provides the appearance and different driving images provide different expressions and head poses. The reenacted face has the same appearance as the source and the same pose-and-expression as the driving. Both the source and driving images are unseen in the training stage.}
		\label{fig1}
	\end{figure*}
	
	\begin{abstract}
		The paper proposes a novel generative adversarial network for one-shot face reenactment, which can animate a single face image to a different pose-and-expression (provided by a driving image) while keeping its original appearance. The core of our network is a novel mechanism called appearance adaptive normalization, which can effectively integrate the appearance information from the input image into our face generator by modulating the feature maps of the generator using the learned adaptive parameters. Furthermore, we specially design a local net to reenact the local facial components (i.e., eyes, nose and mouth) first, which is a much easier task for the network to learn and can in turn provide explicit anchors to guide our face generator to learn the global appearance and pose-and-expression. Extensive quantitative and qualitative experiments demonstrate the significant efficacy of our model compared with prior one-shot methods.
	\end{abstract}

	\section{Introduction}\label{sec:introduction}
	
	In this paper we seek a one-shot face reenactment network, which can animate a single source image to a different pose-and-expression (provided by a driving image) while keeping the source appearance (i.e identity). We start with the perspective that a face image can be divided into two parts, the pose-and-expression and the appearance, which is also adopted by previous work~\cite{OneShotFace2019}. In face reenactment, the transferring of pose-and-expression is relatively easy because the training data can cover most possible poses and expressions. The main challenge of face reenactment is how to preserve the appearances of different identities. This insight motivates us to design a new architecture, which exploits a novel mechanism called the appearance adaptive normalization, to better control the feature maps of the face generator for the awareness of the source appearance. In general, the appearance adaptive normalization can effectively integrate the specific appearance information from the source image into the synthesized image, by modulating the feature maps of the face generator. Especially, the appearance adaptive normalization learns specific adaptive parameters (i.e., mean and variance) from the source image, which are utilized to modulate feature maps in the generator. In this way, the face generator can be better aware of the appearance of the source image and effectively preserve the source appearance.

	The appearance adaptive normalization is inspired by recent adaptive normalization methods~\cite{Huang_2017,park2019semantic}, which perform cross-domain image generation without retraining for a specific domain. This attribute makes adaptive normalization potentially suitable for one-shot face reenactment, in which each identity could be seen as a domain. However, there exists a key challenge to apply these adaptive normalization methods to face reenactment. That is, these existing adaptive normalization methods are all designed to deal with the pixel-aligned image-to-image translation problems. For example, in~\cite{park2019semantic} they propose spatially-adaptive normalization for synthesizing photorealistic images given an input semantic layout. However, in the scenario of face reenactment, the source and driving images are not pixel-aligned. Such pixel mis-alignment makes it difficult to optimize the adaptive normalization layers during training in existing methods. Consequently, the existing methods will yield distorted images after reenactment, and we will show it in the experiments.
	To tackle this challenge, one key insight of our work is that instead of learning individual adaptive parameters for different adaptive normalization layers using independent architectures, we can use a unified network to learn all the adaptive parameters from the source image in a global way. The benefit of such paradigm is, by jointly learning the adaptive parameters, the different adaptive normalization layers can be globally modulated rather than being modulated locally. In this way, we can effectively optimize the adaptive normalization layers and control the feature maps of face generator to keep the source appearance. Specifically, we design a simple but effective skip-connected network to predict the adaptive parameters from the source image, which can explicitly promote the relations within adaptive parameters for different adaptive normalization layers, and thus effectively propagate the appearance information throughout the network during reenacting.

	\begin{figure*}[ht]
		\centering
		\includegraphics[scale=0.55]{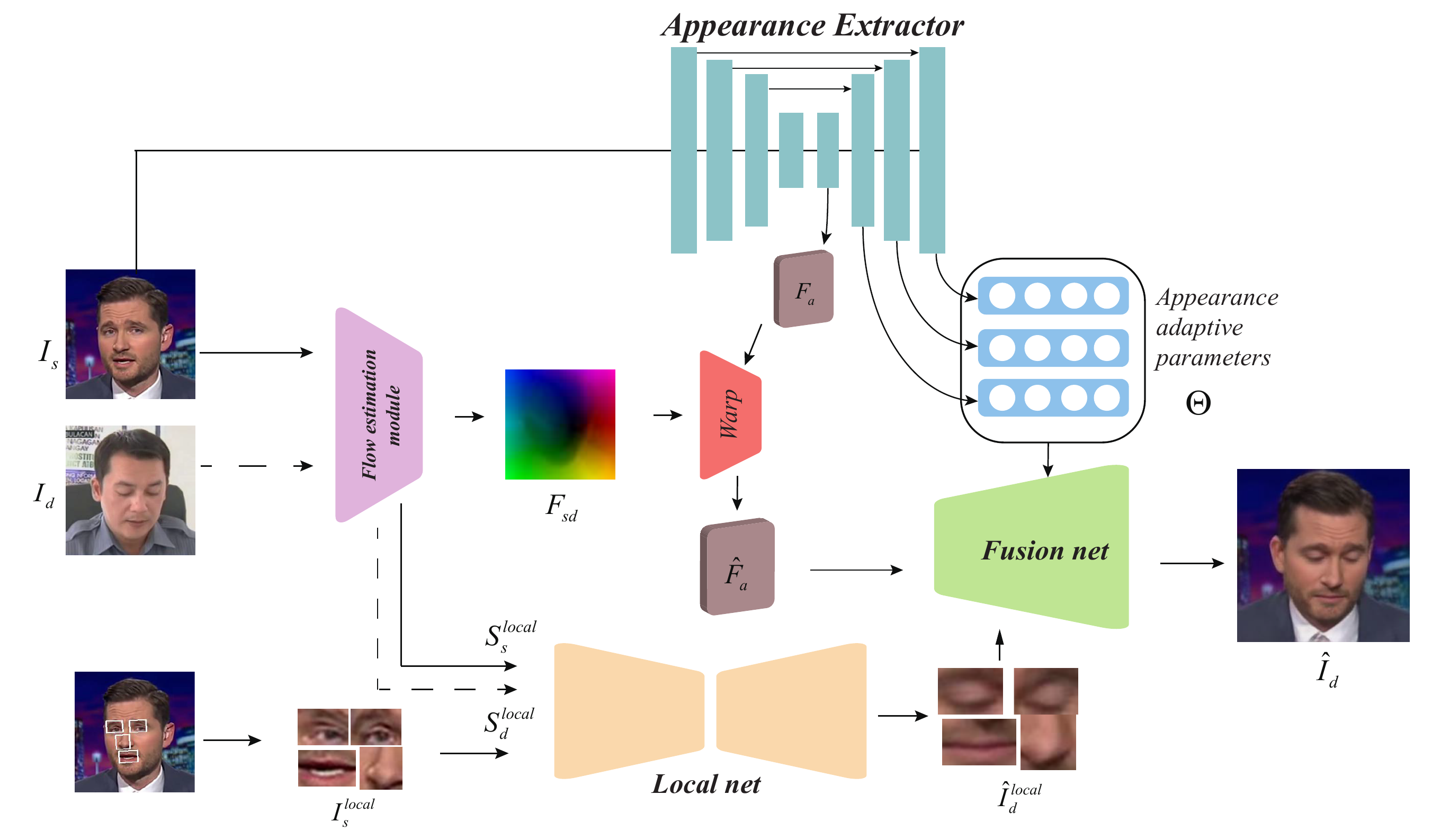}
		\\
		\caption{The architecture of generator of our proposed method.}
		\label{genernator}
	\end{figure*}
	
	We make another key observation that, compared with reenacting the whole faces with largely varying appearances and expressions, reenacting the local facial components (i.e., eyes, nose, and mouth) is a much easier task for the network to learn. It is because the space of appearance and pose-and-expression is significantly reduced for these local regions. To this end, we can learn the reenactment of these local regions first, which can in turn provide explicit anchors to guide our generator to learn the global appearance and pose-and-expression. Especially, the landmarks are utilized to locate the source and target positions of each face component, so the network only needs to learn the reenactment of these components locally. After local reenacting, the synthesized face components are transformed to the target  positions and scales with a similarity transformation and fed to the global generator for the global face synthesis. 
	
	In summary, we propose a novel framework for one-shot face reenactment, which utilizes appearance adaptive normalization to better preserve the appearance during reenacting and local facial region reenactment to guide the global synthesis of the final image. Our model only requires one source image to provide the appearance and one driving image to provide the pose-and-expression, both of which are unseen in the training data. The experiments on a variety of face images demonstrate that our method outperforms the state-of-the-art one-shot methods in both objective and subjective aspects (e.g., photo-realism and appearance preservation).
	
	The main contributions of our work are:
	
	\begin{itemize}
		\item [1)]
		We propose a novel method for one-shot face reenactment, which animates the source face to another pose-and-expression while preserving its original appearance using only one source image. In particular, we propose an appearance adaptive normalization mechanism to better retain the appearance. 
		\item [2)]
		We introduce the reenactment of local facial regions to guide the global synthesis of the final reenacted face. 
		\item [3)]
		Extensive experiments show that our method is able to synthesize reenacted images with both high photo-realism and appearance preservation.
	\end{itemize}
	
	\section{Related Work}
	
	\subsection{Face Reenactment}
	
	Face reenactment is a special conditional face synthesis task that aims to animate a source face image to a pose-and-expression of driving face. Common approaches to face reenactment could be roughly divided into two categories: many-to-one and many-to-many. Many-to-one approaches perform face reenactment for a specific person.
	ReenactGAN~\shortcite{wu2018reenactgan} utilizes CycleGAN~\shortcite{CycleGAN2017} to convert the facial boundary heatmaps between different persons, and hence promote the quality of the result synthesized by an identity-specific decoder. 
	Face2Face~\shortcite{thies2016face2face} animates the facial expression of source video by swapping the source face with the rendered image. 
	The method of \citet{kim2018deep} can synthesize high-resolution and realistic facial images with GAN.
	However, all these methods require a large number of images of the specific identity for training and only reenact the specific identity.
	On the contrary, our method is capable of reenacting any identity given only a single image without the need for retraining or fine-turning. 
	
	To extend face reenactment to unseen identities, some many-to-many methods have been proposed recently. 
	\citet{zakharov2019few} adopt the architecture of BigGAN~\shortcite{brock2018large} and fashional meta-learning, which is capable of synthesizing a personalized talking head with several images, but it requires fine-tuning when a new person is introduced.
	\citet{OneShotFace2019} propose an unsupervised approach to face reenactment, which does not need multiple poses for the same identity. Yet, the face parsing map, an identity-specific feature, is utilized to guide the reenacting, which leads to distorted results when reenacting a different identity.
	\citet{geng2018warp} introduce warp-guided GANs for single-photo facial animation. However, their method needs a photo with frontal pose and neutral expression, while ours does not have this limitation. \cite{pumarola2018ganimation} generates a face guided by action units~\shortcite{friesen1978facial}, which makes it difficult to handle pose changes. 
	X2Face~\shortcite{wiles2018x2face} is able to animate a face under the guidance of pose, expression, and audio, but it can not generate face regions that do not exist in original images.
	MonkeyNet~\shortcite{Siarohin_2019_CVPR} provides a framework for animating general objects. However, the unsupervised keypoints detection may lead to distorted results in the one-shot case.
	MarioNetTe~\shortcite{ha2020marionette} proposes the landmark transformer to preserve the source shape during reenactment, but it does not consider how to retain the source appearance.
	Yao~\etal~\shortcite{Yao_2020} introduce graph covolutional network to learn better optical flow, which helps method to yield better results.
	Different from previous many-to-many methods, our goal is to synthesize a high-quality face image, by learning the appearance adaptive parameters to preserve the source appearance and utilizing the local component synthesis to guide the global face synthesis.

	\subsection{Adaptive Normalization}
	
	The idea of adapting features to different distributions has been successfully applied in a variety of image synthesis tasks~\cite{Huang_2017,park2019semantic}. The adaptive normalization normalizes the feature to zero mean and unit deviation first, and then the normalized feature is denormalized by modulating the feature using the learned mean and standard deviation. In conditional BN~\cite{courville2017modulating}, the fixed categorical images are synthesized using different parameters of the normalization layers for different categories. However, unlike the categorical image generation with fixed categories, the number of identities is unknown in the one-shot face reenactment.
	AdaIN~\cite{Huang_2017} predicts the adaptive parameters for style transfer, which is spatially sharing. However, it is insufficient in controlling the global appearance, since the facial appearance is spatially varying.
	SPADE~\cite{park2019semantic} deploys a spatially varying normalization, which makes it suitable for spatially varying situations. However, SPADE~\cite{park2019semantic} is designed for the pixel-aligned image translation task which uses independent blocks to locally predict the adaptive parameters for different layers.
	In face reenactment, the source and driving images are not pixel-aligned, which makes it difficult to locally optimize the different adaptive normalization layers. Hence, we propose the appearance adaptive normalization mechanism to globally predict adaptive parameters of different layers using a skip-connected network, which better promotes the relations within the adaptive parameters for different layers during transferring.
	
	\section{Methodology}

	For convenience, we denote the images in the dataset as ${I_i^j }_{i=1,…,N_j}^{j=1,…,M}$, where $j$ denotes the identity index and $i$ denotes the image index of identity $j$. $M$ is the number of identities and $N_j$ is the number of images of identity $j$. $S_i^j \in \mathbb{R}^{68 \times H \times W}$ denotes the corresponding heatmaps for the 68 facial landmarks of $I_i^j \in \mathbb{R}^{3 \times H \times W}$, where $H$ and $W$ are the image height and width.
	
	\subsection{Overview}
	
	Our method is a generative adversarial method. We adopt a self-supervised approach to train the network in an end-to-end way, where the driving image $I_d$ has the same identity as $I_s$ in the training stage (i.e., two frames from a video).
	The landmark transformer~\cite{ha2020marionette} is utilized to improve the identity preservation.
	Fig.\ref{genernator} shows the architecture of the proposed generator, which takes as input the source image $I_s$ and the driving image $I_d$. 
	Our generator is composed of 4 sub-nets, and all the 4 sub-nets are jointly trained in an end-to-end way. First, to preserve the source appearance, we send $I_s$ to the appearance extractor to learn the appearance adaptive parameters $\Theta$ as well as the encoded appearance feature $F_a$, as shown at the top of Fig.~\ref{genernator}. Second, to estimate the facial movements from the source image to the driving pose-and expression, the flow estimation module estimates the optical flow $F_{sd}$ from $I_s$ to $I_d$ , which is then utilized to warp the encoded appearance feature, as shown in the middle of Fig.~\ref{genernator}. Third, the local net is deployed to reenact the local facial regions, which provides essential anchors to guide the subsequent synthesis of the whole face, as shown at the bottom of Fig.~\ref{genernator}. Finally, the fusion net fuses the adaptive parameters $\Theta$, the reenacted local face regions $\hat{I}_d^{local}$ and the warped appearance feature $\hat{F}_a$, to synthesize the reenacted face. By modulating the distribution of feature maps in the fusion net using the appearance adaptive parameters, we let $F_{sd}$ determine the pose-and-expression, and $F_a$ and $\Theta$ retain the appearance.

	\subsubsection{Flow Estimation Module}
	
	\begin{figure}[ht]
		\centering
		\includegraphics[scale=0.45]{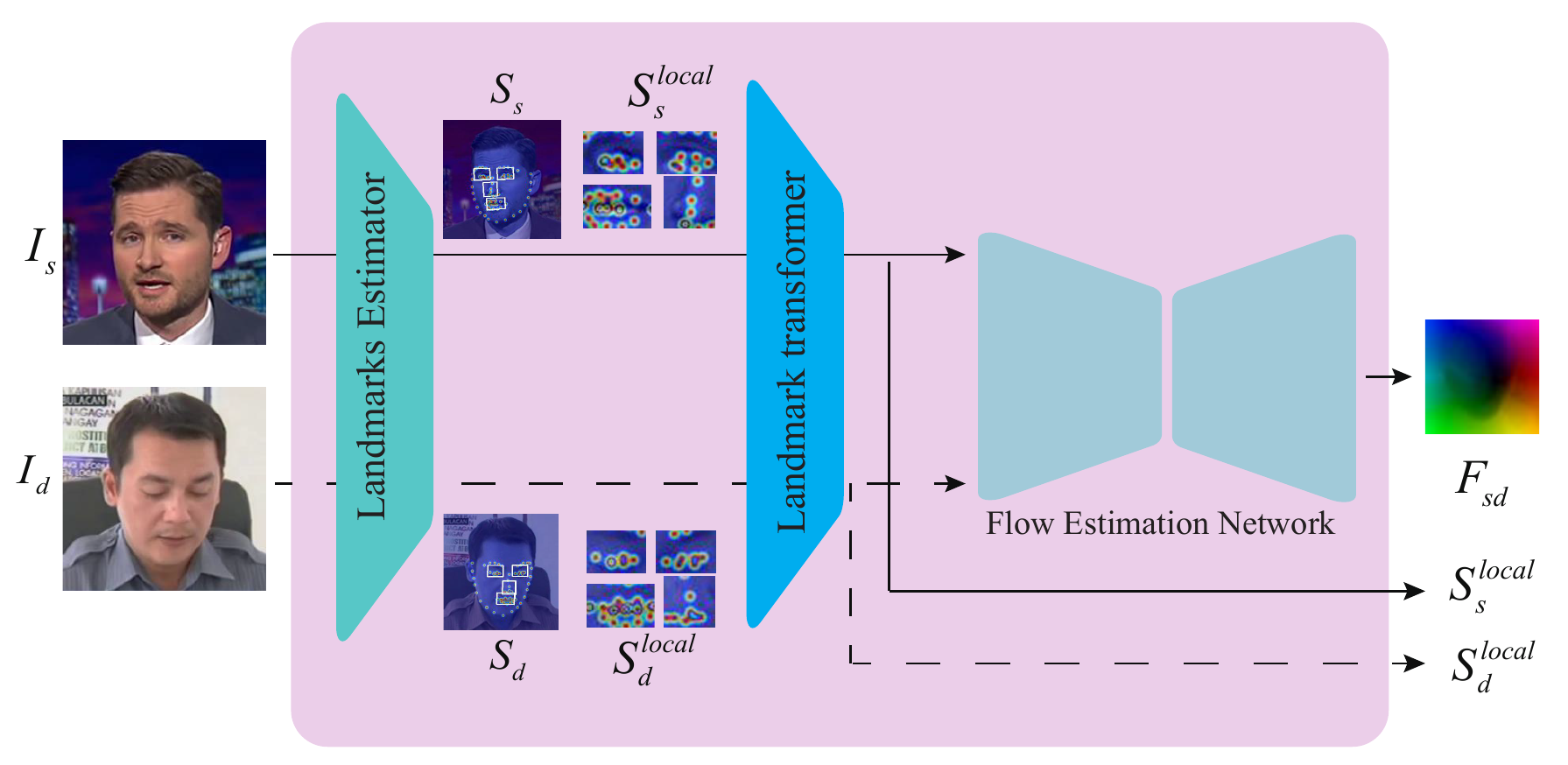}
		\\
		\caption{The procedure of flow estimation module.}
		\label{fig:flow}
	\end{figure}
	
	The procedure of flow estimation module is illustrated in Fig.~\ref{fig:flow}. 
	Firstly, we estimate landmarks for $I_s$ and $I_d$ to obtain the source heatmap $S_s$ and the driving heatmap $S_d$ respectively using OpenFace\cite{amos2016openface}. 
	We then feed $S_s$ and $S_d$ into the flow estimation net (FEN) to produce an optical flow $F_{sd} \in \mathcal{R}^{2\times H \times W}$, representing the motion of pose-and-expression. 
	$F_{sd}$ is then utilized to warp the appearance feature $F_a$. Bilinear sampling is used to sample $F_{sd}$ to the spatial size of $F_a$. The warped $F_a$ is denoted as $\hat{F}_a$, which is subsequently fed into the fusion net to synthesize the final reenacted face. 
	Besides, we also build the heatmaps of local regions for source and driving images based on the landmarks, denoted as $S_s^{local}$ and $S_d^{local}$ respectively.
	The architecture of FEN is an hourglass net~\cite{yang2017stacked}, composed of several convolutional down-sampling and up-sampling layers. 
	Notably, large shape differences between the source identity and the driving identity will lead to severe degradation of the quality of generated images, which is also mentioned by~\cite{wu2018reenactgan}.
	To deal with this issue, we additionally adopt the landmark transformer~\cite{ha2020marionette}, which edits the driving heatmap $S_d$ so that $S_d$ has a shape close to $S_s$. For more details, please refer to ~\cite{ha2020marionette}.
	
	\subsubsection{Local Net}
	
	The local net $G_{local}$ is built with the U-Net structure~\cite{Ronneberger2015UNetCN}. We reenact the left eye, right eye, nose and mouth with 4 independent networks $G_{eyel}$, $G_{eyer}$, $G_{nose}$, and $G_{mouth}$. Each of them is a U-Net with three down-convolution blocks and three up-convolution blocks. The inputs of each local generator are $I_s^{local}$, $S_s^{local}$ and $S_d^{local}$, where \textit{local} refers to the corresponding parts (i.e., left eye, right eye, nose and mouth) on the image and heatmap. The reenacted face local regions serve as anchor regions that can effectively guide the fusion net to synthesize the whole reenacted face.
	
	\subsubsection{Appearance Extractor}
	
	The source image $I_s$ is fed into the appearance extractor $E_a(I_s)$ for predicting the adaptive parameters $\Theta$ and the appearance feature $F_a$. 
	Here $\Theta = \{\theta_i = (\gamma_i,\beta_i), i \in \{1,2,...,N_a\}\}$, where $i$ is the index of the adaptive normalization layer and $N_a$ denotes the number of adaptive normalization layers in the fusion net. 
	For a feature map $F_i\in \mathcal{R}^{c\times h \times w}$ in the fusion net, we have the corresponding $\gamma_i, \beta_i \in \mathcal{R}^{c\times h \times w}$ to modulate it. The encoded source appearance feature $F_a$ is warped to $\hat{F}_a$ using the optical flow $F_{sd}$, and $\Theta$ and $\hat{F}_a$ are fed to the fusion net for face synthesis by controlling the distributions of feature maps. We employ the U-net~\shortcite{Ronneberger2015UNetCN} architecture for the appearance extractor, because the skip-connection in appearance extractor can effectively promote the relations between adaptive parameters.
	
	\subsubsection{Fusion Net}
	
	\begin{figure}[h]
		\centering
		\includegraphics[scale=0.46]{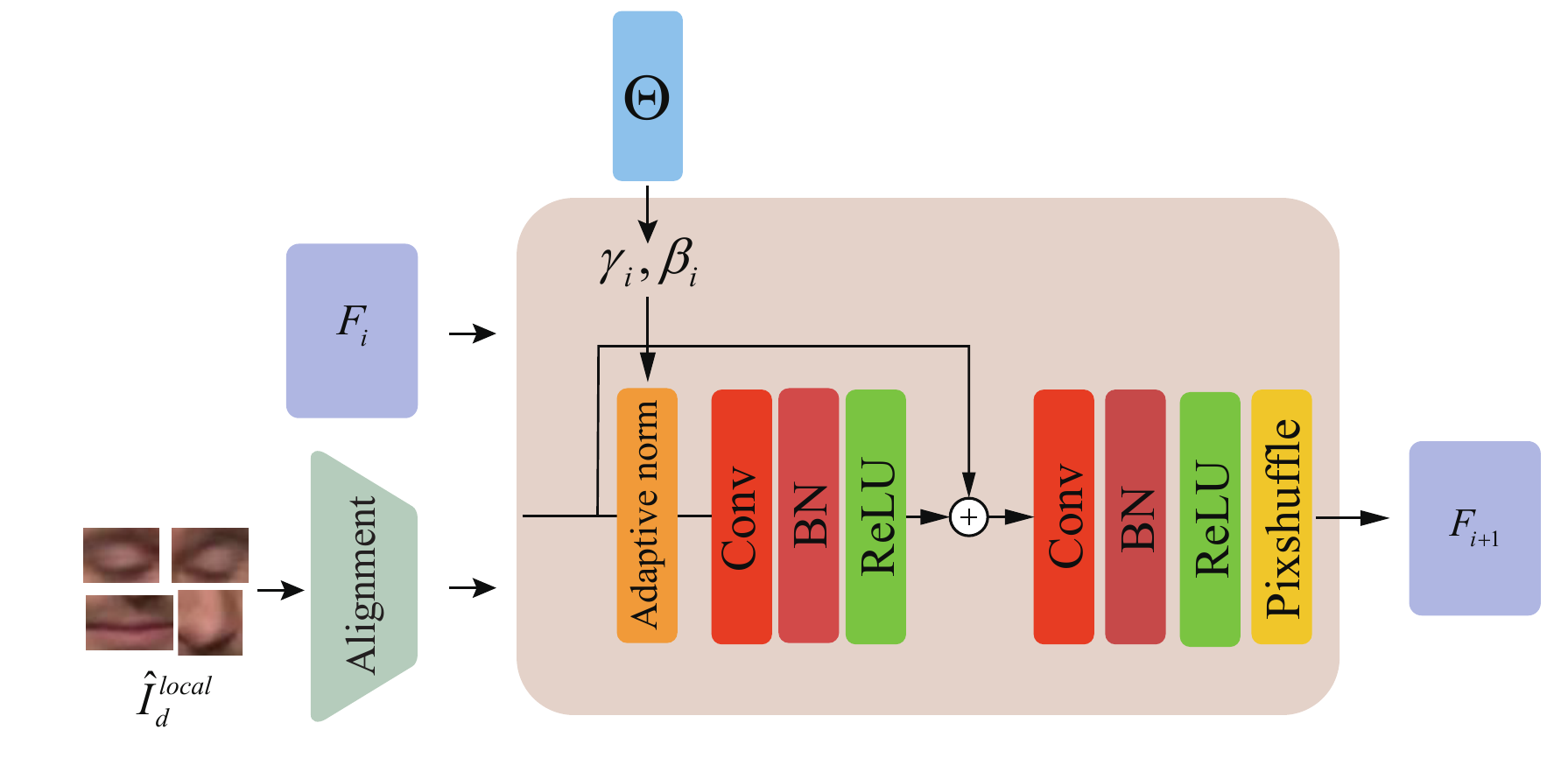}
		\caption{The fusion block of the proposed method.}
		\label{Fusion}
	\end{figure}
	
	The fusion net $ \hat{I}_d = G_f(\hat{I}_d^{local},{\hat{F}_a},\Theta)$ aims to decode the reenacted local regions $I_d^{local}$ and the warped appearance feature $\hat{F}_a$ to a reenacted face image $\hat{I}_d$ under the control of adaptive parameters $\Theta$. $G_f$ is a fully convolutional network, which performs decoding and up-sampling to synthesize the reenacted face. $G_f$ consists of several fusion blocks to adapt the source appearance, followed by several residual-connected convolution layers to produce the final result. 
	The architecture of fusion block is illustrated in Fig.~\ref{Fusion}.
	$F_i$ denotes the input feature map of $i$-th fusion block, $\gamma_i$ and $\beta_i$ denote the $i$-th adaptive parameters and $FB_i$ denotes the $i$-th fusion block. 
	Before fed into the fusion block, the reenacted local regions $\hat{I}_d^{local}$ are similarly transformed to the target scale-and-position. In this way, the aligned face regions provide explicit anchors to the generator. 
	These aligned $\hat{I}_d^{local}$ are then resized to the same spatial size as $F_{i}$ using bilinear interpolation. At last, $F_{i}$ and $\hat{I}_d^{local}$ are concatenated along the channel axis and fed into next block of $G_f$. 
	In this way, the formulation of fusion block can be written as:
	\begin{equation}
		F_{i+1} = FB_i([F_{i},\hat{I}_d^{local}],\gamma_i,\beta_i).
	\end{equation}
	The core of our fusion net is the appearance adaptive normalization mechanism. Specifically, the feature map is channel-wisely normalized by
	\begin{equation}
		\mu_c^i  = \frac{1}{NH^iW^i}\sum_{n,h,w} F_{n,c,h,w}^i ,
	\end{equation}
	\begin{equation}
		\sigma_c^i  =\sqrt{\frac{1}{NH^iW^i}\sum_{n,h,w} [(F_{n,c,h,w}^i)^2 - (\mu_c^i)^2]},
	\end{equation}
	where $F_{n,c,h,w}^i$ is the feature map value before normalization, and $\mu_c^i$ and $\sigma_c^i$ are the mean and standard deviation of the feature map in channel $c$. The index of the normalized layer is denoted as $i$. Notably, the denormalization in adaptive normalization is element-wise, where the normalized feature map is denormalized by
	\begin{equation}
		\gamma_{c,h,w}^i \frac{F_{n,c,h,w}^i -\mu_c^i }{\sigma_c^i} + \beta_{c,h,w}^i\label{(eq:5)}.
	\end{equation}
	Here $\gamma_{c,h,w}^i$ and $\beta_{c,h,w}^i$ are the scale and bias learned by the appearance extractor from $I_s$. Besides, instead of using the transposed convolutional layer or the bilinear up-sampling layer followed by a convolutional layer to expand the feature-map~\cite{Isola_2017,Wang_2018}, we adopt the pixel-shuffle~\cite{shi2016real} to upscale the feature map.
	
	\subsection{Discriminator}
	
	There are two discriminators in our method, a discriminator $D_L$ to discriminate whether the reenacted image and the driving heatmap are matched (pose-and-expression consistency) and a discriminator $D_I$ to discriminate whether the source and reenacted image share the same identity (appearance consistency). $D_L$ takes $\hat{I}_d$ and $S_d$ as input, while $D_I$ takes $\hat{I}_d$ and $I_s$ as input. $\hat{I}_d$ is concatenated with $S_d$ or $I_s$ along the channel axis, before being fed into $D_L$ or $D_I$ respectively.
	To generate a sharp and realistic-looking image, the discriminators should have a large receptive field~\cite{Wang_2018}. In our method, instead of using a deeper network with larger convolutional kernels, we use a multi-scale discriminator~\cite{Wang_2018} which can improve the global consistency of generated images in multiple scales.

	\begin{figure*}
		\centering
		\scalebox{0.62}[0.62]{
			\includegraphics[scale=0.65]{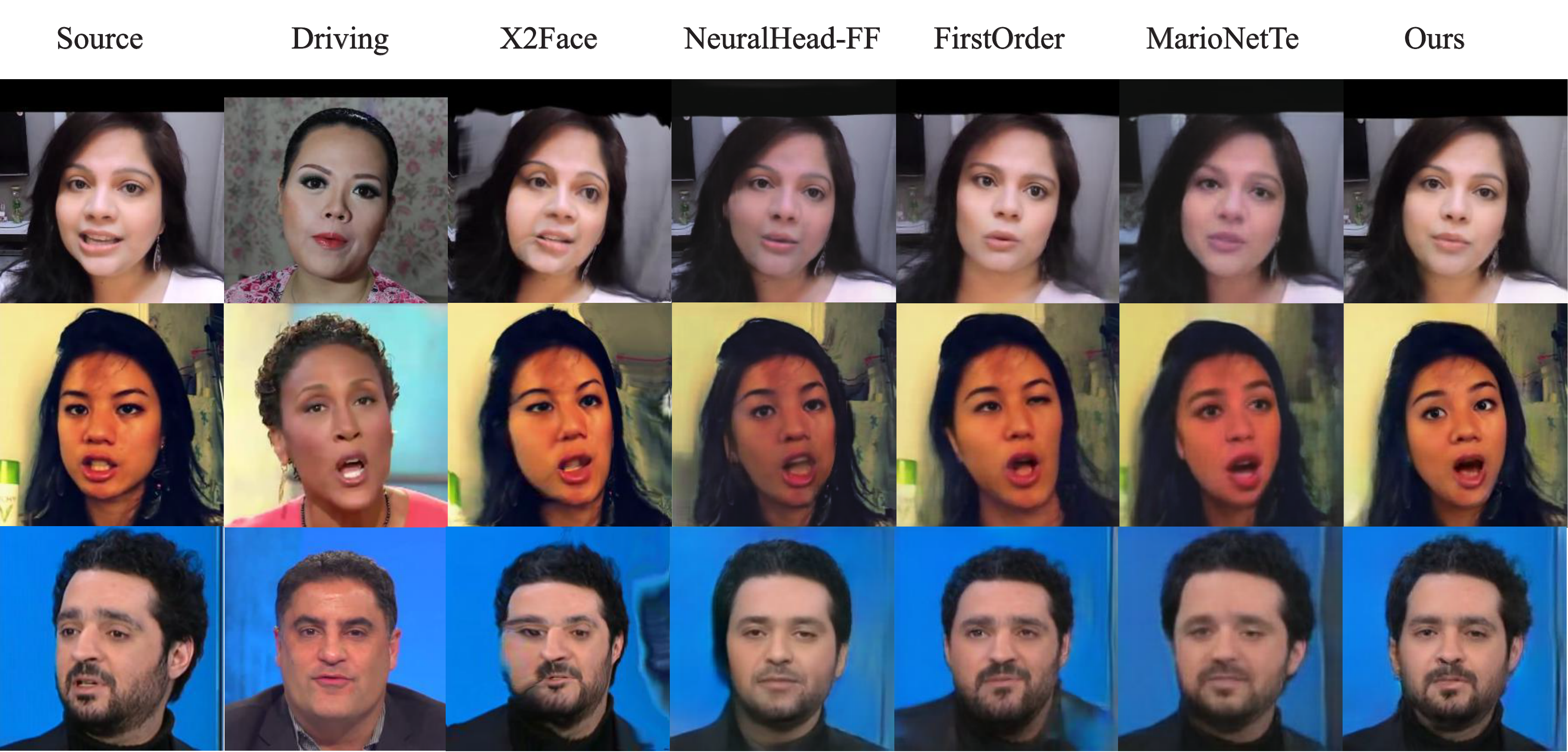}}
		\caption{Qualitative comparison with state-of-the-art one-shot methods.
			Our proposed method generates more natural-looking and sharp results compared to previous methods.}
		\label{methods_compare}
	\end{figure*}
	
	\subsection{Loss Function}
	
	The total loss function is defined as:
	\begin{equation}
		\begin{split}
			L_{total} = \arg \min_G \max_{D_L,D_I} \lambda_{GAN}L_{GAN}
			+ \lambda_c L_c \\
			+  \lambda_{local} L_{local}.,
		\end{split}
	\end{equation}
	where $L_c$ denotes the content loss, $L_{GAN}$ denotes the adversarial loss and $L_{local}$ denotes local region loss.
	The adversarial loss is the GAN loss for $D_L$ and $D_I$:
	\begin{equation}
		\begin{split}
			L_{GAN} = \mathbb{E}_{I_s,\hat{I}_d,S_d} [ \log D_L(I_d,S_d) + \log(1-D_L(\hat{I}_d,S_d)) ] \\
			+ \left.\mathbb{E}_{I_s,\hat{I}_d,I_d} [ \log D_I(I_s,I_d) + \log(1-D_I(I_s,\hat{I}_d,I_d))]\right. .
		\end{split}
	\end{equation}
	The content loss is defined as:
	\begin{equation}
		L_c = L_1(I_d,\hat{I}_d) + L_{per}(I_d,\hat{I}_d) \label{con:inventoryflow},
	\end{equation}
	where $L_1(I_d,\hat{I}_d)$ is the pixel-wise L1 loss, measuring the pixel distance between the generated image and the ground-truth image. $L_{per}(I_d,\hat{I}_d)$ is the perceptual loss~\cite{johnson2016perceptual}, which has been shown to be useful for the task of image generation~\cite{ledig2017photo}. We make use of the pre-trained VGG~\cite{simonyan2014very} to compute the perceptual loss, and $L_{per}$ is written as:
	\begin{equation}
		L_{per}(I_d,\hat{I}_d) = \mathbb{E}_{i \in X} [||\Phi_i(I_d)-\Phi_i(\hat{I}_d)||_1],
	\end{equation}
	where $X$ represents the layers we use in VGG and $\Phi_i(x)$ denotes the feature map of the $i$-th layer in $X$.
	
	The local region loss penalizes the perceptual differences between the reenacted local regions and the local regions on the ground-truth and is defined as:
	\begin{equation}
		\begin{split}
			L_{local} &= L_{per}(I_{eyel}, \hat{I}_{eyel}) + L_{per}(I_{mouth}, \hat{I}_{mouth}) \\
			&+ L_{per}(I_{nose}, \hat{I}_{nose}) + L_{per}(I_{eyer},\hat{I}_{eyer}).
		\end{split}
	\end{equation}

	\section{Experiments}

	\subsection{Implementation}
	
	The learning rate for the generator and discriminator are set to $2e^{-5}$ and $1e^{-5}$ respectively. We use Adam~\cite{kingma2014adam} as the optimizer. Spectral Normalization~\cite{miyato2018spectral} is utilized for each convolution layer in the generator. We set $\lambda_{GAN}=10$, $\lambda_{c}=5$ and $\lambda_{local} = 5$ in the loss function. The Gaussian kernel variance of heatmaps is 3.
	
	\subsection{Datasets and Metrics}
	
	Both the FaceForensics++~\cite{roessler2019faceforensicspp}, VoxCeleb1\cite{nagrani2017voxceleb} and Celeb-DF~\cite{li2020celeb} datasets are used for quantitative and qualitative evaluation. The OpenFace~\cite{amos2016openface} is utilized to detect the face and extract facial landmarks.
	Following the work of MarionNetTe\shortcite{ha2020marionette}, we adopt the following metrics to quantitatively evaluate the reenacted faces of different methods. 
	We evaluate the identity preservation by calculating the cosine similarity (CSIM) of identity vectors between the source image and the generated image. The identity vectors are extracted by the pre-trained state-of-the-art face recognition networks~\cite{deng2019arcface}.
	To inspect the model's capability of properly reenacting the pose and expression of driving image, we calculate PRMSE~\cite{ha2020marionette} and AUCON~\cite{ha2020marionette} between the generated image and the driving image to measure the reenacted pose and expression respectively.

	\subsection{Quantitative and Qualitative Comparison}
	
%
%
%
%
%
%
%
%
%
%
%
%
%
%
%
		\begin{table}[h]
		\centering
		\scalebox{1}[1]{
			\begin{tabular}{cccccc}
				\toprule
				Model &CSIM$\uparrow$  &PRMSE$\downarrow$ &AUCON$\uparrow$\\
				\midrule
				
				\multicolumn{4}{c}{VoxCeleb1~\shortcite{nagrani2017voxceleb}}\\
				\midrule
				
				X2face & 0.689  &\underline{3.26} &0.813\\
				
				NeuralHead-FF & 0.229  &3.76 &0.791\\
				
				MarioNETte & 0.755 &\textbf{3.13} &0.825\\
				
				FirstOrder &0.813  &3.79 &\textbf{0.886}\\
				
				Ours &\textbf{0.823}  &\underline{3.26} &\underline{0.831} \\
				
				\midrule
				\multicolumn{4}{c}{Celeb-DF~\shortcite{li2020celeb}}\\
				\midrule
				
				X2face &0.676  &4.10 &0.679 \\
				
				NeuralHead-FF &0.511   &6.09 &0.747 \\
				
				MarioNETte  &0.650   &3.98 &0.714 \\
				
				FirstOrder &\underline{0.687}  &\underline{3.15} &\textbf{0.839} \\
				
				Ours &\textbf{0.753} &\textbf{3.12} &\underline{0.751} \\
				
				\bottomrule
		\end{tabular}}
	\caption{Quantitative comparison in the self-reenactment setting. Up/down arrows correspond to higher/lower values for better performance. Bold and underlined numbers represent the best and the second-best values of each metric respectively.}
	\label{tab:Quantitative_Comparison}
	\end{table}

	\begin{table}[h]
		\centering
		\scalebox{1}{
			\begin{tabular}{cccc}
				\toprule 
				Model &CSIM$\uparrow$ &PRMSE$\downarrow$ &AUCON$\uparrow$\\
				
				\midrule 
				\multicolumn{4}{c}{Faceforensics++~\shortcite{roessler2019faceforensicspp}}\\
				\midrule 
				
				X2face &0.604 &9.80 &0.697\\
				
				NeuralHead-FF &0.381 &6.82 &\underline{0.730} \\
				
				MarioNETte &\underline{0.620} &7.68 &0.710\\
				
				FirstOrder &0.614 &\textbf{6.62} &\textbf{0.734} \\
				
				Ours &\textbf{0.658}  &\underline{7.04} &0.706 \\
				
				\midrule 
				\multicolumn{4}{c}{Celeb-DF~\shortcite{li2020celeb}}\\
				\midrule 
				
				X2face &0.400 &6.52 &0.400\\
				
				NeuralHead-FF &0.352 &8.30 &0.480 \\
				
				MarioNETte &\underline{0.460} &\underline{5.16} &\textbf{0.662}\\
				
				FirstOrder & 0.432 &6.10 &0.500 \\
				
				Ours &\textbf{0.463} &\textbf{5.10} &\underline{0.660}   \\
				
				\bottomrule 
		\end{tabular}}
	\caption{Quantitative comparison of reenacting a different identity.}
	\label{tab:DifferenceID_Comparison}
	\end{table}

	Table~\ref{tab:Quantitative_Comparison} lists the quantitative comparisons with existing one-shot reenactment methods when reenacting the same identity, and Table~\ref{tab:DifferenceID_Comparison} reports the evaluation results when reenacting a different identity.
	It is worth mentioning that the method that, following \cite{ha2020marionette}, we re-implement \cite{zakharov2019few} using only the feed-forward network in the one-shot setting.
	Differ from other competitors, FirstOrder~\shortcite{Siarohin_2019_NeurIPS} require two driving image to perform the relative motion transfer, one image provide the initial driving pose-and-expression and another one to provides the target driving pose-and-expression.
	We use the source image to provide the initial driving pose-and-expression when reenacting the same identity to perform the relative motion transfer, and the absolute motion transfer is adopted when reenacting different identities as the initial driving image is lacked for all competitors.
	
	Notably, the results show that our method outperforms other methods in many metrics, demonstrating our method can synthesize highly realistic faces while effectively retaining the source appearance and faithfully reenacting the pose-and-expression.
	Fig.~\ref{methods_compare} illustrates typical qualitative examples, all of which are randomly selected from the testing set.
	We can see that X2face~\shortcite{wiles2018x2face} is unable to generate face regions that do not exist in the source images, so it may result in large artifacts. As the state of art, MarioNETte~\shortcite{ha2020marionette} can effectively preserve the source shape, but there may still be some appearance artifacts in some regions. 
	Our method fixes this issue by introducing the appearance adaptive normalization and local region reenacting.

	\begin{figure}[h]
		\centering
		\scalebox{0.8}[0.8]{\includegraphics[scale=0.20]{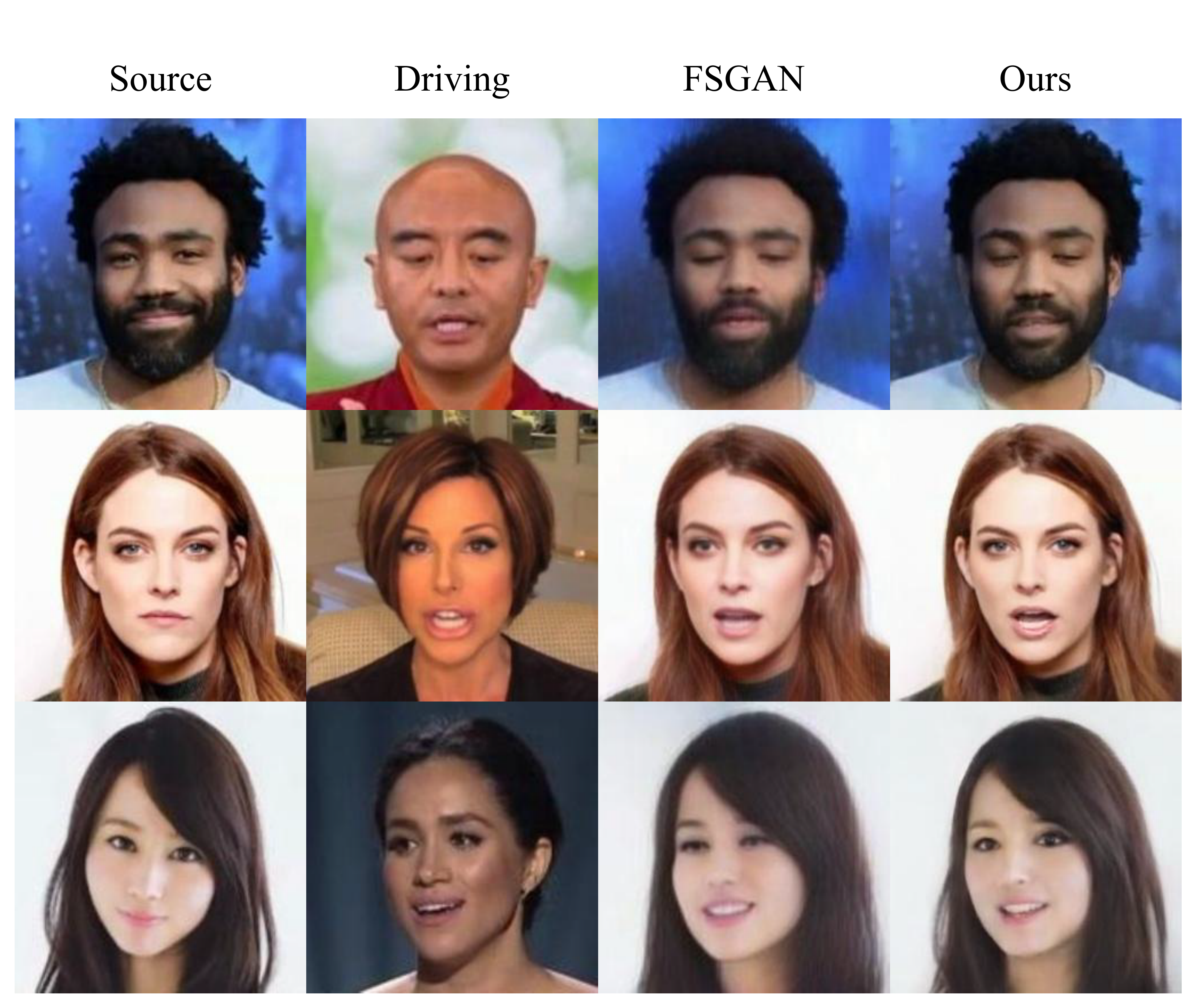} }
		\caption{Comparison of our method with FSGAN\shortcite{nirkin2019fsgan}, source andn driving images are cited from FSGAN\shortcite{nirkin2019fsgan}}
		\label{fig:cmp_fsgan}
	\end{figure}
	
	\begin{figure}[h]
		\centering
		\scalebox{0.7}[0.7]{\includegraphics[scale=0.23]{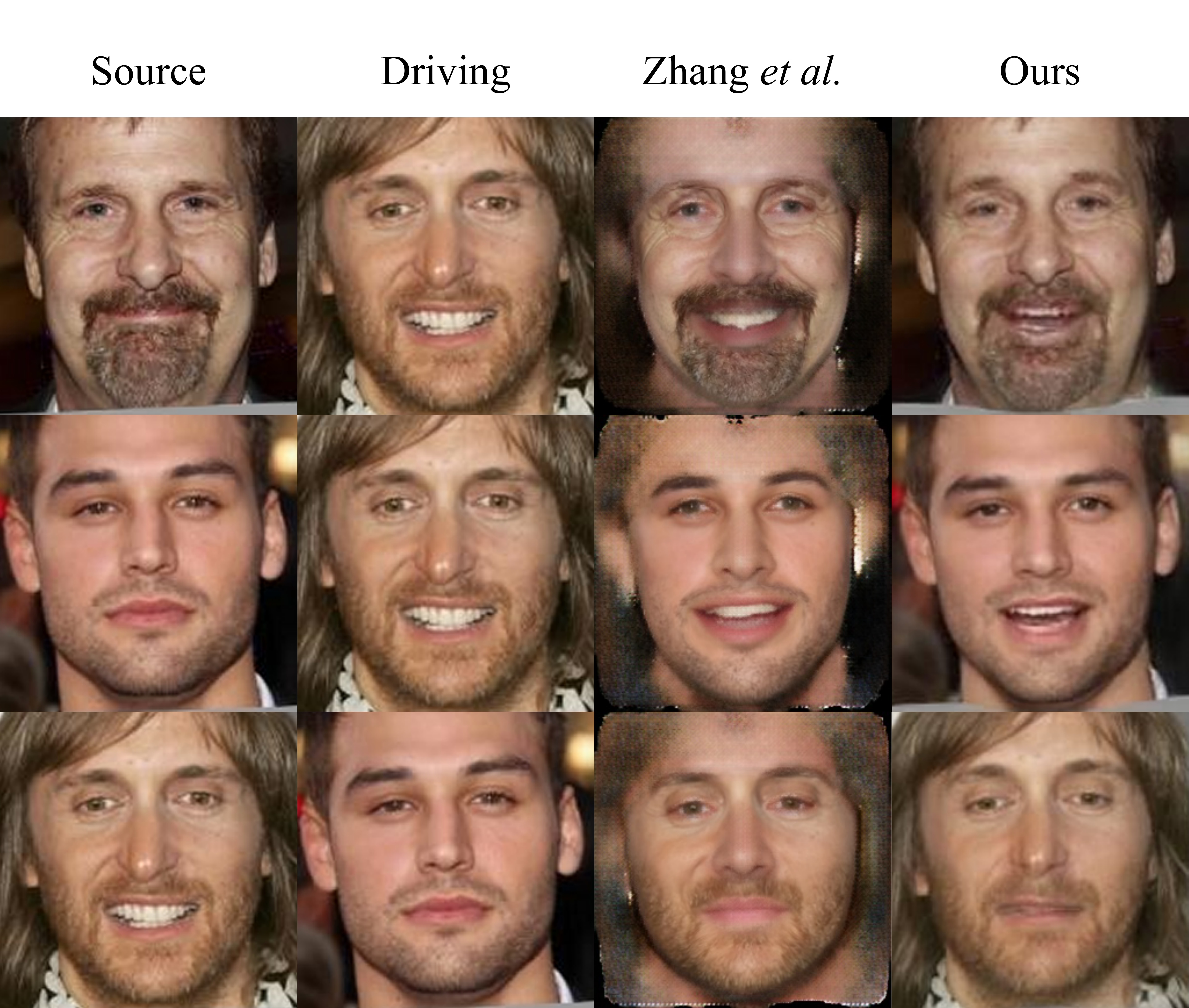} }
		\caption{Comparison of our method with Zhang~\etal\shortcite{OneShotFace2019}, source andn driving images are cited from Zhang~\etal\shortcite{OneShotFace2019}.}
		\label{fig:cmp_oneshot}
		
	\end{figure}

	We also qualitatively compare our method with recently proposed methods of \citet{OneShotFace2019} and FSGAN\shortcite{nirkin2019fsgan}, demonstrated in Fig.~\ref{fig:cmp_fsgan} and Fig.~\ref{fig:cmp_oneshot}. We can observe blurriness and color-inconsistency in the results of FSGAN\shortcite{nirkin2019fsgan}. Also the images synthesized by \citet{OneShotFace2019} have distorted face shapes and artifacts in boundaries, because \citet{OneShotFace2019} utilize the face parsing map, which is an identity-specific feature, to guide the reenacting. On the contrary, with the help of appearance adaptive normalization and local region reenacting, our method can achieve more detailed and natural-looking results.
	
	\subsection{Ablation Study}
	
	\begin{figure}[h]
		\centering
		\scalebox{0.7}[0.7]{
			\includegraphics[scale=0.52]{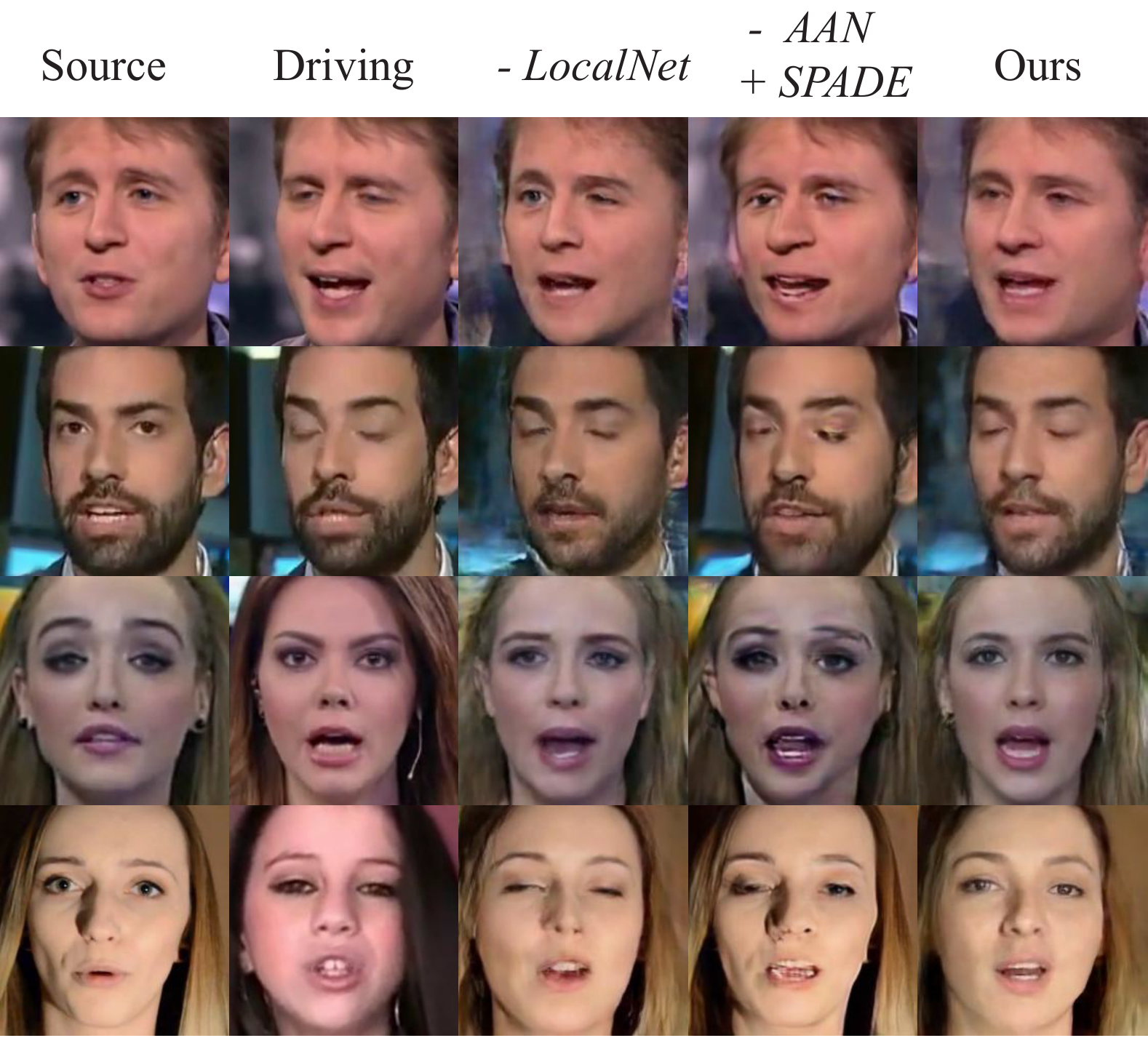}}
		\caption{Qualitative results of the ablation study. Our full model leads to better results than other variants.}
		\label{ablation_result}
	\end{figure}

	\begin{table}[h]
		\centering
		\begin{tabular}{ccccccc}
			\toprule
			\multicolumn{4}{c}{Model}        &CSIM$\uparrow$ &PRMSE$\downarrow$ &AUCON$\uparrow$ \\
			
			\midrule
			\multicolumn{4}{c}{- local net} &0.615 &7.293 &0.698 \\
			\multicolumn{4}{c}{- AAN + SPADE} &0.558 &11.030 &0.660\\
			\multicolumn{4}{c}{Ours} &\textbf{0.658}  &\textbf{7.04} &\textbf{0.706} \\
			\bottomrule
		\end{tabular}
	\caption{Quantitative ablation study for reenacting a different identity on the Faceforensics++.}
	\label{tab:DifferenceID_Ablation}
	\end{table}
	
	To better evaluate the key components within our network, we perform the ablation study by evaluating the following variants of our method:
	\begin{itemize}
		\item $- LocalNet$. The local net is excluded from the full model.
		\item $-AAN+SPADE$. To validate the effectiveness of appearance adaptive normalization, we use the spatially-adaptive normalization to replace it, and all the other components are the same as our model.
	\end{itemize}
	
	The qualitative results are illustrated in Fig.~\ref{ablation_result} and quantitative results are listed in Table~\ref{tab:DifferenceID_Ablation}. 
	We can see that our full model presents the most realistic and natural-looking results.
	The local net can help reduce the pose-and-expression error, as it explicitly provides anchors for local face regions to guide the reenacting. 
	The appearance adaptive normalization can effectively improve image quality and reduce artifacts by globally modulating the appearance features. 
	Compared to the spatially-adaptive normalization~\shortcite{park2019semantic}, our appearance adaptive normalization can better preserve the source appearance and leads to more realistic results. It validates our appearance adaptive normalization is more suitable for face reenactment.
	
	\subsection{Conclusion and Future Work}
	
	In the paper, we propose a novel method to deal with the challenging problem of one-shot face reenactment. Our network deploys a novel mechanism called appearance adaptive normalization to effectively integrate the source appearance information into our face generator, so that the reenacted face image can better preserve the same appearance as the source image. Besides, we design a local net to reenact the local facial components first, which can in turn guide the global synthesis of face appearance and pose-and-expression. Compared to previous methods, our network exhibits superior performance in different metrics.
	
	\section{Acknowledgments}
	We thank anonymous reviewers for their valuable comments. This work is supported by National Key R\&D Program of China (2018YFB1004300), NSF China (No. 61772462,  No. U1736217) and the 100 Talents Program of Zhejiang University.
	
	\bibliography{REF.bib}

\end{document}